%% file: main.tex
\documentclass[conference]{IEEEtran}
\IEEEoverridecommandlockouts
\usepackage{cite}

\usepackage{amsmath,amssymb,amsfonts}
\usepackage{algorithmic}
\usepackage{graphicx}
\usepackage[dvipsnames]{xcolor}
\usepackage{textcomp}
\usepackage{xcolor}
\usepackage{lipsum}  
\usepackage{enumitem}
\usepackage{multirow}
\usepackage{booktabs}
\usepackage{colortbl}

\newcolumntype{s}{>{\columncolor[gray]{.90}[.5\tabcolsep]}c}

\input{macros}

\def\BibTeX{{\rm B\kern-.05em{\sc i\kern-.025em b}\kern-.08em
    T\kern-.1667em\lower.7ex\hbox{E}\kern-.125emX}}

\definecolor{myBlue}{RGB}{66, 119, 182}
\definecolor{eqnBlue}{HTML}{8FAADC} %
\definecolor{eqnRed}{HTML}{C45812} %
\definecolor{myRed}{RGB}{186, 28, 48}
\definecolor{citecolor}{HTML}{D73F09} %

\usepackage[pagebackref=true,breaklinks=true,urlcolor=citecolor,colorlinks,citecolor=citecolor,bookmarks=false,linkcolor=citecolor]{hyperref}

\newcommand{\ours}[0]{\texttt{PCWM}}
\begin{document}
\IEEEsettopmargin{t}{72pt}
\title{Point Cloud Models Improve \\ Visual Robustness in Robotic Learners}
 \author{\IEEEauthorblockN{Skand Peri $^1$}
 \and
 \IEEEauthorblockN{Iain Lee $^2$}
 \and
 \IEEEauthorblockN{Chanho Kim $^1$}
 \and
 \IEEEauthorblockN{Li Fuxin $^1$}
 \and
 \IEEEauthorblockN{Tucker Hermans $^{2,3}$}
 \and
 \IEEEauthorblockN{Stefan Lee $^1$}
 }

\maketitle

\footnotetext[1]{Oregon State University $^2$University of Utah $^3$ NVIDIA}

\begin{abstract}
\input{sections/01_abstract}
\end{abstract}

\begin{IEEEkeywords}
point cloud world model, model-based reinforcement learning, vision-based robot control, robustness
\end{IEEEkeywords}

\section{Introduction}
\input{sections/02_intro}

\section{Related Work}
\input{sections/03_related_works}

\section{Point Cloud World Models}
\input{sections/04_method}

\section{Experimental Setup}

\input{sections/05_experiments}

\section{Results}
\input{sections/06_results}

\section{Discussion \& Limitations}
\label{sec:analysis}
\input{sections/07_discussion}

\section{Conclusion}
\input{sections/08_conclusion}

\section{Acknowledgements}
\input{sections/09_acknowledgements}

\bibliographystyle{ieeetr}
\bibliography{references}

\newpage
\section{Appendix}
\subsection{Hyperparameters}
\input{sections/appendix/01_hyperparameters}

\subsection{Environment Details}
\input{sections/appendix/02_environment_details}

\end{document}

%% file: macros.tex
\renewcommand{\arraystretch}{1.25}
\usepackage{xspace}

\newcommand{\xhdr}[1]{\vspace{3pt}\noindent\textbf{#1}}
\newcommand{\uhdr}[1]{\vspace{3pt}\noindent\underline{\smash{#1}}}

\newcommand{\tabref}[1]{Tab.~\ref{#1}\xspace}
\newcommand{\figref}[1]{Fig.~\ref{#1}\xspace}
\newcommand{\secref}[1]{Sec.~\ref{#1}\xspace}

\newcommand{\eg}{e.g.\xspace}
\newcommand{\ie}{i.e.\xspace}

\newcommand{\rgbd}{RGB-D\xspace}
\newcommand{\rgbdwm}{RGBD-WM\xspace}

%% file: sections/01_abstract.tex
Visual control policies can encounter significant performance degradation when visual conditions like lighting or camera position differ from those seen during training -- often exhibiting sharp declines in capability even for minor differences. In this work, we examine robustness to a suite of these types of visual changes for RGB-D and point cloud based visual control policies. To perform these experiments on both model-free and model-based reinforcement learners, we introduce a novel Point Cloud World Model (PCWM) and point cloud based control policies. Our experiments show that policies that explicitly encode point clouds are significantly more robust than their RGB-D counterparts. Further, we find our proposed PCWM significantly outperforms prior works in terms of sample efficiency during training. Taken together, these results suggest reasoning about the 3D scene through point clouds can improve performance, reduce learning time, and increase robustness for robotic learners. \newline
\textbf{Project Webpage}: \href{https://pvskand.github.io/projects/PCWM}{https://pvskand.github.io/projects/PCWM}
\vspace{5pt}

%% file: sections/02_intro.tex
To broaden the application and deployment of robot manipulators in the world, we must extend their understanding of and ability to operate in unstructured environments~\cite{Bhattacharjee2018TowardsRF}. However, the dynamics of such environments contain significant uncertainty. Furthermore, robots can typically only sense these environments through partial observations. Modeling every aspect of the world ``in the wild" is thus intractable. Owing to this, planning under such situations can be prohibitively expensive especially in unseen scenes when novel objects are introduced. Hence, to endow manipulators to act in complex scenarios with only partial view sensing information, recent works have relied on learning based robotic control \cite{finnDeepVisuomotor, Nair2022R3MAU, Hansen2022pre}.

However, learning-based robot control policies that rely on imagery as input can exhibit significant performance degradations when visual conditions like lighting, camera position, or object textures differ from those seen during training \cite{Xie2023DecomposingTG}. This lack of robustness is a hurdle for in-the-wild deployment and has prompted the extensive study of data augmentation \cite{yarats2021image, Yarats2021MasteringVC, Hansen2021StabilizingDQ} and pretraining \cite{Nair2022R3MAU, radosavovic2023real, Parisi2022TheUE} techniques in visual policy learning. In this work, we examine the question of policy robustness from the perspective of visual input representation -- finding policies that encode observations as XYZ-RGB point clouds rather than \rgbd images demonstrate greater robustness.

\begin{figure}[t]
\hspace{0.1in}\includegraphics[width=0.9\linewidth]{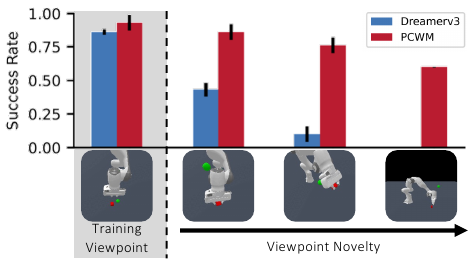}
  \vspace{-8pt}
  \caption{\textit{Motivating Example.} We compare  \textbf{\textcolor{myBlue}{DreamerV3}}, a state-of-the-art RL model that is trained on \rgbd inputs with our \textbf{\textcolor{myRed}{Point Cloud World Model (\ours{})}} on a simple task of lifting a cube. We find the point clouds are significantly more robust to viewpoint changes compared to \rgbd.}
  \label{fig:00_intro}
  \vspace{-11pt}

\end{figure}

To illustrate this phenomenon, we examine a simple control task in \figref{fig:00_intro} where a robotic arm must lift a red cube from a table up to a green goal point. We trained state-of-the-art model-based reinforcement learning (RL) policies \cite{hafner2023dreamerv3} for this task with \rgbd input (denoted \textcolor{myBlue}{DreamerV3}) and point cloud input (denoted \textcolor{myRed}{PCWM}). When tested on novel viewpoints, we observe that the success rate of DreamerV3 drops by half even for a slight variation and fails completely on more significant changes. In contrast, the \ours{}'s performance decays much more slowly. We find similar trends in our larger suite of experiments later in this paper. At first glance, the reasons for this are unclear. XYZ-RGB point clouds and \rgbd images contain much of the same information. 
How then can we account for such a difference between these policies?

We hypothesize this difference comes from \emph{how} these modalities are typically encoded. In standard practice, RGB-D inputs are encoded with convolutional networks -- simply treating the depth information as a fourth image channel. Under these architectures, convolutional kernels aggregate features based on closeness in the 2D pixel space, even when neighboring pixels may have vastly different depths and thus correspond to different parts of the scene. This could lead to features between far away objects being averaged together, leading to worse performance.  In contrast, point cloud representations allow the XYZ coordinates to directly serve as features, enabling networks to learn geometric invariances and equivariances such as those with respect to rotation and scaling~\cite{li2023improving} -- which are functions of the XYZ coordinates.%

To study this phenomenon in robot control settings, we develop a suite of point cloud-based control policies and train them with state-of-the-art model-free and model-based reinforcement learning algorithms. For model-based, we propose a first-of-its-kind point cloud-based world model (\ours{}). We train these models on a suite of robot control tasks and examine generalization to out-of-distribution camera viewpoints, field-of-view, lighting conditions, and distractor objects. We find point cloud-based models to be significantly more robust than their RGB-D counterparts -- even maintaining performance under large shifts in visual conditions. Further, we find our \ours{} model-based framework  achieves better sample efficiency and higher task performance than its RGB-D model-based counterpart \cite{hafner2023dreamerv3} on several manipulation tasks.

\xhdr{Contributions.} We summarize our main contributions:
\begin{itemize}[leftmargin=10pt]
    \item We study robustness to changes in viewpoint, field-of-view, lighting, and distractor objects for RGB-D and point cloud-based visual control policies.
    \item We propose Point Cloud World Models (\ours{}s), a model-based reinforcement learning framework based  on partial point clouds. We show gains in sample efficiency and robustness over comparable RGB-D models.
    \item Beyond increased robustness, we show \ours{}s adapt more quickly when finetuned in new environments with significant differences in visual conditions.
\end{itemize}

%% file: sections/03_related_works.tex
\noindent\textbf{Point Clouds in RL.}
Visual policy learning has seen significant progress in game playing \cite{Mnih2015HumanlevelCT, Hasselt2015DeepRL}, robotic and dexterous manipulation \cite{Zhu-RSS-18, chen2022towards, Handa2023DeXtremeTO}, and locomotion tasks \cite{Yarats2021MasteringVC, hafner2019dreamer, Kaiser2019ModelBasedRL}. Most of this work leverages RGB(-D) imagery, hence explicit consideration for 3D representation learning has been limited \cite{singh2023selfsupervised, thomason2022language, Ze2022VisualRL}. 
Several recent works have proposed \emph{model-free} policies that learn from partial point clouds \cite{liu2022frame, dexpoint, ling2023efficacy} -- demonstrating that the rich 3D information in point clouds can improve sample efficiency in interactive robotic tasks. We extend this body of work by (1) introducing a novel \emph{model-based} RL framework for point clouds (\ours{}) %
and (2) demonstrating that point clouds offer increased visual robustness for both model-based and model-free policies.
Recently, GROOT \cite{Zhu2023LearningGM} showed how point clouds can be robust to environment changes in the context of imitation learning, however, we focus on agents trained with RL policies.

\xhdr{Robustness in RL.}
Prior work has demonstrated that vision-based policies learned from RGB(-D) input can have poor generalization to new visual conditions \cite{Cobbe2018QuantifyingGI, Song2019ObservationalOI, Lyle2022LearningDA, Yang2023WhatIE, krantz2022sim2sim, krantz2023navigating}. These include changes due to new task instances, differences in object textures or lighting, novel viewpoints, or a combination of these induced by sim-to-sim or sim-to-real transfer. Inspired by work in computer vision, data augmentation \cite{Kostrikov2020ImageAI, Yarats2019ImprovingSE, Yarats2021MasteringVC} and representation pretraining \cite{Nair2022R3MAU, Xiao2022MaskedVP, Parisi2022TheUE} techniques have been employed to ameliorate this lack of robustness. These methods require careful design of image augmentations or laborious curation of diverse pretraining datasets to improve generalization \cite{robonet, roboturk, ebert2021bridge}. While these techniques have demonstrated positive impacts, visual control policies for robotics can still exhibit a significant generalization gap \cite{Xie2023DecomposingTG}. In this work, we study the role of input representation in policy robustness. Our findings suggest that point cloud-based policies can be robust to viewpoints, lighting conditions and addition of new objects in the scene even \textit{without} any of the above techniques.

\xhdr{Model-based RL.} One technique in sequential decision making is to learn a model of the environment \cite{sutton1991dyna} and use it for planning \cite{walsh2010integrating, pineau2003point, pets, shrestha2020deepaveragers} or policy learning \cite{hafner2019dreamer, hafner2020dreamerv2, hafner2023dreamerv3, Rajeswar2023MasterURLB}. In the case of high dimensional inputs such as images, a popular approach is to learn the environment dynamics in a compact latent space that is supervised using rewards \cite{Hansen2022TemporalDL, hansen2022modem} and image reconstruction \cite{hafner2019planet, op3, ha2018world, finnDeepVisuomotor}. Such model-based RL agents \cite{hafner2019dreamer, hafner2020dreamerv2, hafner2023dreamerv3, Rajeswar2023MasterURLB} have showcased higher sample efficiency compared to analogous model-free policies.  However, these works have focused on settings where observations are RGB images or privileged state information such as the location of scene objects. We propose the first point cloud world model and investigate its sample efficiency and robustness. 

\xhdr{Point Cloud Dynamics.} Prior work has proposed variants of graph neural networks \cite{47094} to learn dynamics with point clouds \cite{li2018learning, Sanchez-Gonzalez20}.
While these approaches can model realistic collision dynamics, they require point-to-point correspondences between frames. When deploying these models as part of a planning system in the real-world, prior work has applied mesh reconstruction on point clouds obtained by either multiple cameras \cite{shi2023robocook} or a single RGB-D camera \cite{shi2022robocraft}, which could be prone to errors for novel objects. While a dynamics model that takes partial point clouds as input was proposed \cite{Huang-icra2023-graph-relations}, it requires 6-DoF object poses for its supervision and was not tested within an RL framework.  In this work, we propose the first point cloud dynamics model that enables world model training in RL by directly operating on partial point clouds and using only the reward signal for its supervision.

%% file: sections/04_method.tex
Our model-based reinforcement learning approach for point clouds consists of two learned components -- a world model that simulates the effects of actions (\secref{sec:model}) and a policy learned in this simulated environment that maps states to actions (\secref{sec:policy}). As in prior work for high-dimensional inputs \cite{hafner2019planet, hafner2019dreamer, hafner2020dreamerv2, hafner2023dreamerv3}, we consider a latent world model that simulates the world in a learned lower-dimensional space. %

\xhdr{Problem Formulation.} We pose our problem as an infinite-horizon Partially Observable Markov Decision Process (POMDP) \cite{Cassandra1994ActingOI} defined by a tuple $(\mathcal{S}, \mathcal{A}, \mathcal{T}, \mathcal{R}, \mathcal{O}, \gamma, \mathcal{\rho}_0)$. $\mathcal{S}$ represents the state space with a complete scene point cloud, which is not accessible to the agent. The observation space, $\mathcal{O} \in \mathbb{R}^{N \times 6}$ denotes partial point cloud observations with $N$ points featurized with position $(x, y, z)$ and color $(r, g, b)$. $\mathcal{A} \in \mathbb{R}^m$ is an $m$-dimensional continuous action space, $\mathcal{T}: \mathcal{O} \times \mathcal{A} \rightarrow \mathcal{O}$ is the transition function, $\mathcal{R}: \mathcal{O} \rightarrow \mathbb{R}$ is the reward function, $\gamma \in [0, 1)$ is the discount factor and $\mathcal{\rho}_0$ denotes the initial state distribution. The goal of the agent is to learn a policy $\pi: \mathcal{O} \rightarrow \mathcal{A}$ that maximizes the expected sum of discounted rewards; $\max_{\pi}\mathbb{E}_\pi[\sum_{t=1}^{\infty} \gamma^t \mathcal{R}(s_t)]$.

\subsection{World Model}
\label{sec:model}
We base our world model on the Recurrent State-Space Model (RSSM) framework \cite{hafner2019planet} which learns a recurrent world model with a $d$-dimensional latent variable $z$. The RSSM model is derived from an evidence lower bound on the likelihood of an observation sequence $o_{1:T}$ given actions $a_{1:T}$. This results in a loss composed of two components: a reconstruction term measuring how well observations (and rewards) can be predicted from the latent representation and a KL divergence term keeping predicted latent states near corresponding real observation encodings.

For point cloud observations, designing the reconstruction task is non-trivial -- irregular point densities may bias the loss function to denser regions and jointly predicting the structure and featurization of a point cloud from a latent vector is challenging. To sidestep these issues, we follow \cite{Hansen2022TemporalDL, Oh2017VPN} by dropping observation reconstruction and relying only on multi-step reward prediction and the KL term for supervision.

\begin{figure}[t]
\begin{center}
  \includegraphics[width=0.875\linewidth]{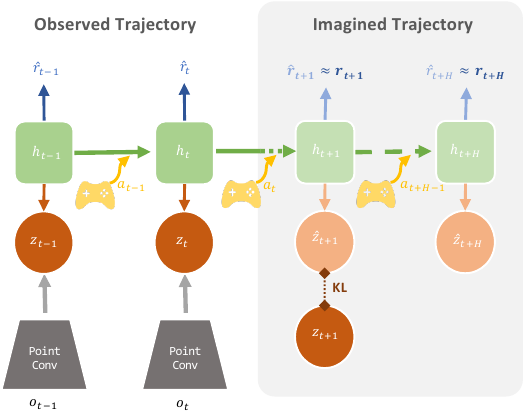}
  \caption{\textbf{\ours{} training}: Given a sequence of $T$ partial point cloud observations $o_{1:T}$, we encode them using a PointConv encoder. For each timestep $t$, we compute a posterior stochastic latent $z_t$ using an encoding of $o_t$ and hidden state $h_t$ that encodes the history. The hidden state is further used to compute the prior latent $\hat{z}_t$ which is used to predict multi-step rewards over a horizon $H$ providing supervision for the world model alongside a KL-loss for temporal consistency See \secref{sec:model}.}
  \label{fig:00_pcwm}
  \vspace{-27pt}
\end{center}
\end{figure}

More concretely, our world model shown in \figref{fig:00_pcwm} is parameterized by $\phi$ and consists of the following components:
\begin{equation}
\begin{array}{lll}
    \text{Representation:} && z_{t} \sim q_{\phi}(h_t, o_{t})\\
    \text{Recurrent Model:} && h_t = f_{\phi}(z_{t-1}, h_{t-1}, a_{t-1})\\
    \text{Dynamics:} && \hat{z}_{t} \sim p_{\phi}(\hat{z}_t | h_t)\\
    \text{Reward:} && \hat{r}_{t} \sim p_{\phi}(\hat{r_t} | h_t, z_t)\\
    \text{Continuation Predictor:} && \hat{c}_{t} \sim p_{\phi}(\hat{c_t} | h_t, z_t)
\end{array}
\end{equation}
where we use $\sim$ to denote the sampling operation. The continuation flag $c_t \in \{0, 1\}$ indicates whether the episode has ended. Except for the input encoder network within $q_\phi$, we retain architectural choices from \cite{hafner2023dreamerv3} for the other components.

Given a partial point cloud $o_t \in \mathbb{R}^{N \times 6}$ with $N$ points, we encode it using a series of PointConv layers \cite{wu2019pointconv} to obtain an embedding $e_t \in \mathbb{R}^{n \times d}$, where $n$ is the number of points in the downsampled point cloud with each point consisting of a $d$-dimensional feature. We then aggregate the features of $n$ points in the point cloud latent space to obtain the $d$-dimensional feature $\texttt{agg}(e_t)$, where \texttt{agg} is an aggregation function that is used to predict the latent $z$. 

Like TD-MPC \cite{Hansen2022TemporalDL} and VPN \cite{Oh2017VPN}, we find that supervising rewards by rolling out each latent in the future for $H$ steps helps learn a better world model and leads to better performance.  Along with this, we simultaneously train the continuation predictor $c_t$ using binary cross entropy loss.
Since $z_t$ is predicted using the input point cloud ($o_t$) and for dynamics model rollouts we do not have access to $o_t$, we employ a KL loss term to ensure that posterior prediction, $z_t$ and prior prediction $\hat{z_t}$ are close. Hence, this way, $\hat{z_t}$ can be used for accurate rollouts to train the policy.
The overall world model training objective can thus be written as follows\\[0pt] 
\begin{equation}
    \begin{aligned}
     L_{\phi} = \sum_{t=1}^T &\overbrace{p_\phi(r_{t}, c_t \mid h_{t}, z_{t})\rule{0pt}{1.25\baselineskip}}^{\text{\textbf{current-step loss}}} + ~~\overbrace{\left( \sum_{i=1}^H p_\phi(\hat{r}_{t+i} \mid h_{t+i}, \hat{z}_{t+i}) \right)}^{\textcolor{eqnBlue}{\text{\textbf{multi-step reward loss}}}}\\[3pt]
     &~+ ~ \underbrace{\rule[-0.5\baselineskip]{0pt}{0\baselineskip}\text{KL}\left(~q_\phi(z_t \mid h_t, o_t) \mid\mid p_\phi(\hat{z_t} \mid h_t)~\right)}_{\textcolor{eqnRed}{\textbf{one-step temporal consistency}}} 
     \end{aligned}
\end{equation}
\vspace{-32pt}
\subsection{Policy Learning}
\label{sec:policy}
For the policy, we adopt the Actor-Critic framework \cite{A2C} similar to DreamerV3 \cite{hafner2023dreamerv3}, which consists of a \textit{Critic} network that predicts the value at a given state and an \textit{Actor} that predicts the action distribution given a state.
\begin{equation}
\begin{array}{lll}
    \text{Actor network:} && a_{t} \sim \pi_{\psi}(a_t | z_t)\\
    \text{Critic network:} && v_\psi(z_t) \approx \mathbb{E}_{q(.|z_t)} [\sum_{\tau=t}^{t+H}(\gamma^{\tau-t}r_{\tau})]\\
\end{array}
\end{equation}
The critic is learned by discrete regression \cite{Imani2018ImprovingRP, hafner2023dreamerv3} using generalized $\lambda$-targets \cite{Schulman2015HighDimensionalCC}. We train the actor network to maximize the value function via dynamics backpropagation \cite{hafner2019dreamer}, updating actor parameters using the gradients computed through the world model.
Further, we use symlog predictions \cite{hafner2023dreamerv3} for the reward predictor and the critic. Symlog is helpful in dealing with environments with varying reward scales across different tasks. The overall framework alternates between the world model training, policy training, and data collection using the most recent policy.

%% file: sections/05_experiments.tex
\begin{figure*}[t]
  \centering
  \includegraphics[width=0.90\linewidth]{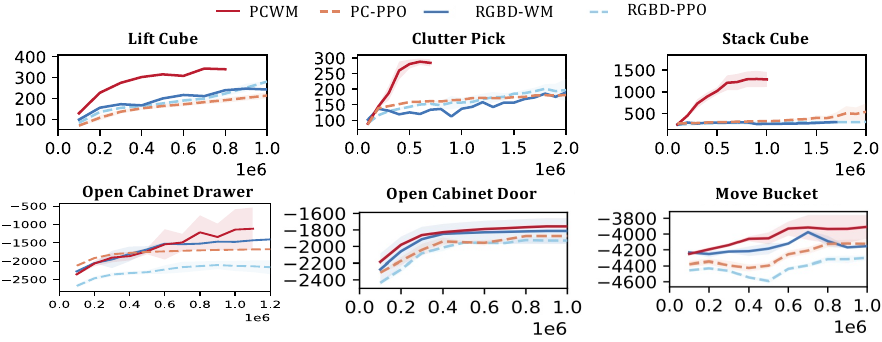}
  \vspace{-11pt}
  \caption{\emph{Task performance:} We report training curves for six manipulation tasks. Our proposed \ours{} either matches or outperforms baselines in all settings -- demonstrating strong sample efficiency gains in several tasks. \ours{} is truncated after achieving task success for Pick \& Place tasks (top row).}
  \label{fig:01_downstream_tasks}
  \vspace{-19pt}
\end{figure*}

\xhdr{Environments.}
We conduct our experiments on the ManiSkill2 \cite{gu2023maniskill2} benchmark on a simulated 7-DoF Franka Panda robotic arm with a parallel gripper. We consider Maniskill2 as it is built on top of the photorealistic physics simulator Sapien \cite{Xiang_2020_SAPIEN} where several works have shown Sim2Real transfer \cite{dexpoint, bao2023dexart, Zhu2023ContactArtL3}. We consider two representative manipulation tasks -- (i) \textit{Pick {\&} Place} and (ii) \textit{Mobile Manipulation}. 

For Pick \& Place, we consider \texttt{LiftCube} (lifting a single cube) and \texttt{StackCube} (stacking one cube on top of another). Additionally, we add a \texttt{ClutteredLiftCube} task, where the goal of the agent is to pick the (unique) red cube among a number of distractor cubes in the scene. For these tasks, we consider an 8-$d$ action space consisting of delta position of arm joints (7) and gripper distance (1). 
For Mobile Manipulation, we consider \texttt{OpenCabinetDrawer} (opening drawer of a cabinet), \texttt{OpenCabinetDoor} (opening a cabinet door) and \texttt{MoveBucket} (moving a bucket from ground to a platform situated at a certain height). For the first two tasks, the agent has a 12-$d$ action space involving manipulation (8) and navigation (4). The latter task is bimanual, adding 8 more dimensions to control a second arm. %
 For more details, we refer the readers to the original papers \cite{mu2021maniskill, gu2023maniskill2}.

\xhdr{Baselines.}
Beyond \ours{}s, we consider representative methods for the other three  \{model-based, model-free\} $\times$ \{\rgbd, point cloud (PC)\} settings. For model-based \rgbd, we modify a stable PyTorch implementation of DreamerV3 \cite{dreamerv3torch} to include depth reconstruction and denote this model as \textbf{\rgbdwm}. For PC and \rgbd model-free policies, we take policy architectures and representation encoders from our corresponding model-based approaches and train them directly from real environment experience using PPO \cite{Schulman2017ProximalPO} -- denoting these models as \textbf{PC-PPO} and \textbf{RGBD-PPO} respectively.

\xhdr{Implementation Details.}
In this section, we discuss several design choices and hyperparameters of our model.%

\uhdr{Point Cloud Encoding.} Using known camera intrinsics and extrinsics, we first transform the point clouds to world coordinates. Then, we use 4 PointConv \cite{wu2019pointconv} layers with a downsampling factor of 2 to encode the input point cloud into $e_t \in \mathbb{R}^{n \times d}$ with $n{=}64$ and $d{=}256$. We aggregate (\texttt{agg}) the point cloud latent using mean pooling to obtain a 256-$d$ representation, which we found to work well across all the tasks. See Sec. \ref{sec:analysis} for discussion of encoder choice. 

\uhdr{Point Pruning.}
Similar to \cite{ling2023efficacy}, we found it helpful to prune distant or task-irrelevant (\eg floor) points in the scene. %
 This pruning could be realized in practice by depth-based clipping, background removal, or object segmentation-based filtering \cite{Huang-icra2023-graph-relations}.
After pruning, we apply farthest point sampling to generate 1024 points for Pick \& Place tasks and 2048 points for Mobile Manipulation tasks. The higher point resolution for the latter set of tasks is to ensure that key parts of the scene such as the door or drawer handle are represented.

\begin{table*}[t]
\begin{center}
\caption{Average reward in original (grey) and visually perturbed settings computed from 25 episodes in each perturbed condition (see \secref{sec:peturb}) and standard deviations across 3 random seeds. \ours{}s achieve higher reward and are more robust to visual changes.}
\label{tab:robust}
\renewcommand*{\arraystretch}{1.15}
\setlength{\tabcolsep}{5pt}
\resizebox{0.925\linewidth}{!}{
\begin{tabular}{l   s c c c   s c c c}
\toprule
 & \multicolumn{4}{c}{\ours{} (Ours)} & \multicolumn{4}{c}{\rgbdwm}\\
\cmidrule(lr){2-5} \cmidrule(lr){6-9}
Task & Original & Viewpoint & Field of View & Lighting & Original & Viewpoint & Field of View & Lighting\\
\midrule
\small Lift Cube & 325 $\pm$ 20 & 280 $\pm$ 50 & 257 $\pm$ 33 & 259 $\pm$ 31 & 305 $\pm$ 13 & 73 $\pm$ 98 & 112 $\pm$ 137 & 126 $\pm$ 11\\
\small Clutter Pick & 358 $\pm$ 29 & 286 $\pm$ 71 & 246 $\pm$ 64 & 349 $\pm$ 27 & 329 $\pm$ 47 & 85 $\pm$ 39 & 30 $\pm$ 13 & 242 $\pm$ 98\\
\small Stack Cube & 1721 $\pm$ 283 & 1269 $\pm$ 412 & 1006 $\pm$ 343 & 1465 $\pm$ 143 & 251 $\pm$ 12 & 193 $\pm$ 59 & 202 $\pm$ 23 & 213 $\pm$ 29\\
\small Open Cabinet Drawer & -500 $\pm$ 32 & -647 $\pm$ 59 & -631 $\pm$ 48 & -549 $\pm$ 43 & -1410 $\pm$ 29 & -2460 $\pm$ 132 & -1782 $\pm$ 238 & -1638 $\pm$ 126\\
\small Open Cabinet Door & -1726 $\pm$ 48 & -1972 $\pm$ 177 & -1983 $\pm$ 56 & -1794 $\pm$ 53 & -1925 $\pm$ 17 & -2303 $\pm$ 396 & -2120 $\pm$ 194 & -2120 $\pm$ 194\\
\small Move Bucket & -3632 $\pm$ 85 & -3901 $\pm$ 135 & -3881 $\pm$ 47 & -3681 $\pm$ 29 & -4168 $\pm$ 89 & -4572 $\pm$ 21 & -4419 $\pm$ 39 & -4276 $\pm$ 55\\
\bottomrule
\end{tabular}}
\end{center}

\vspace{-27pt}
\end{table*}

\uhdr{World Model and Policy Training.}
We pretrain the world model for 1000 steps on 10 random trajectories before starting policy training. For world model training, we uniformly sample sequences of length 64 with a batch size of 8 from the replay buffer. The deterministic state $h$ is 256 dimensional and the continuous stochastic state $z$ is 32 dimensional. The reward and continuation prediction heads are 2 layer MLPs with sigmoid linear unit (SiLU) \cite{SiLU} activation and layer normalization \cite{Ba2016LayerN}. We jointly train the multi-step reward and continuation prediction losses with $H=5$ and the dynamics consistency (KL) loss with an Adam optimizer \cite{kingma2014adam} with a learning rate of 0.0001. For policy training, we rollout using the world model for 15 timesteps from each of the $64$ states of the sampled trajectory. The actor and the value heads are also 2 layer MLPs with SiLU activation and LayerNorm trained with Adam optimizer with a learning rate of $3\mathrm{e}{-5}$.

%% file: sections/06_results.tex
This section is divided into the following claims and supporting evidence from our experimental results.

\subsection{Point Cloud World Models (\ours{}s) can be more sample efficient learners than analogous RGB-D models.} 
\label{sec:efficient}

We show reward curves over the course of training for our six tasks in \figref{fig:01_downstream_tasks}. In all tasks, the proposed \ours{} matches or exceeds the performance of the baseline methods -- including the model-based \rgbdwm \cite{hafner2020dreamerv2}. Strikingly, \ours{}s learn considerably faster in the Pick \& Place style tasks (top row). For example, on \texttt{Clutter Pick} (top middle), \ours{} achieves task success in under 1 million interactions whereas the other methods fail to do so after 2 million. This trend is more pronounced in the \texttt{StackCube} task (top right).
We attribute this gain in efficiency to \ours{}'s ability to reason with explicit 3D representations.
For model free methods, we observe that PC-PPO outperforms RGBD-PPO as well. 

For  \texttt{OpenCabinetDoor} and \texttt{MoveBucket}, \rgbdwm and \ours{} achieved similar performances. We hypothesize that the model-based policy training is a dominant factor in these cases as opposed to the input representation.
The mobile manipulation tasks tend to be more complex, involving navigation to the target object in all three and bimanual coordination for \texttt{MoveBucket}. While all models achieve $>75\%$ success rates for \texttt{OpenCabinetDrawer}, we find they struggle on \texttt{OpenCabinetDoor} and \texttt{MoveBucket}, indicating the need for more interaction to achieve task success. 

\subsection{Point cloud-based policies are more robust to changes in visual conditions than analogous RGB-D policies.}
\label{sec:peturb}

\begin{figure*}[t] 
\begin{center}
  \hspace{0.0in}%
  \includegraphics[width=0.85\linewidth]{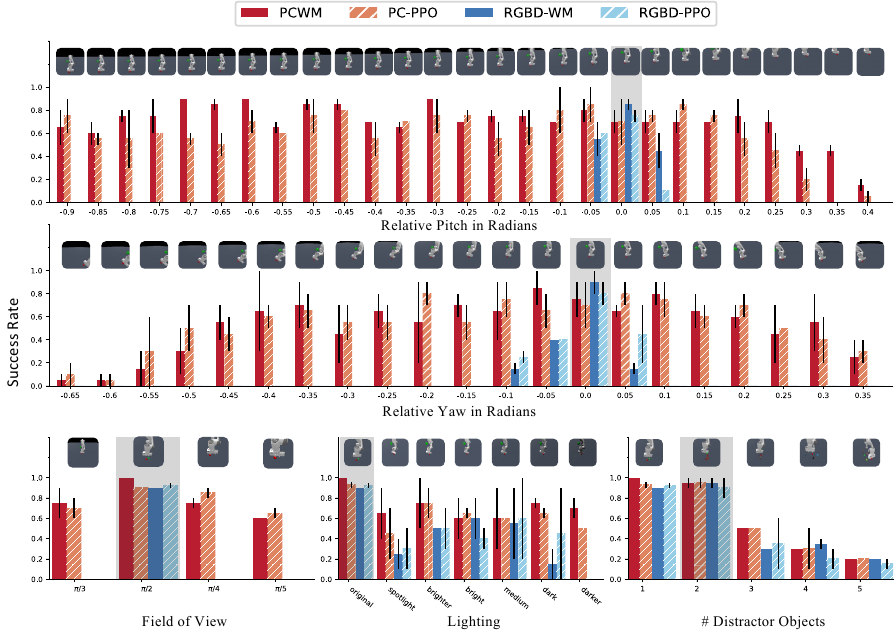}
  \vspace{-8pt}
  \caption{Fine-grained robustness analysis for the \texttt{Clutter Pick} task. Example frames from each condition are shown above policy performance plots. Grey shaded backgrounds indicate the original training environment. We find RGB-D models generalize poorly to new viewpoints or FoV in this setting.}
  \label{fig:02_robustness}
  \vspace{-11pt}
  \end{center}
\end{figure*}

The models in the previous discussion were all trained in \emph{single, fixed imaging conditions} -- \ie with a fixed camera viewpoint, field of view, and scene lighting. Here, we examine their performance when these conditions are systematically varied. Note this set of experiments does not involve any further policy training. Below we describe these variations:
\begin{itemize}[leftmargin=10pt]
    \item \emph{Viewpoint.} We alter either camera pitch or yaw  by 0.05 radian increments through ranges that keep the task objects and manipulators in frame. We select -0.9 to 0.4 radians for yaw and -0.65 to 0.35 for pitch for a total of 42 conditions.
    \item \emph{Field of View.} We vary the field of view of the camera at three discrete levels $\{\frac{\pi}{2}, \frac{\pi}{4}, \frac{\pi}{5}\}$ yielding 3 conditions.
    \item \emph{Lighting.} We consider 6 lighting conditions -- varying ambient illumination through 5 stages from bright to dark and adding a yellow spotlight focused on the table.
\end{itemize}
These conditions are visualized in \figref{fig:02_robustness} for the \texttt{Clutter Pick} task and average rewards across these conditions for all tasks are shown in \tabref{tab:robust} for \ours{} and \rgbdwm. We take the model with best return to compute the results across 3 different seeds. Given their lower overall performance, we do not include the model-free methods in this comparison.

Across settings, we find \ours{} policies achieve significantly better performance than those from \rgbdwm. However, in many tasks this difference in performance was also evident in the original unperturbed setting due to \ours{}'s increased sample efficiency. Focusing on the \texttt{LiftCube} setting where both methods achieve similar task performance in the original environment, we still observe significant differences in performance in perturbed conditions. For example, \ours{} achieves only 6\% less average reward across viewpoint changes compared to a \textbf{76\% reduction} for \rgbdwm.%

\uhdr{Finer-grained Analysis.} The above analysis aggregates over a range of conditions to provide a general sense of policy robustness. To examine this more closely, we take \texttt{Clutter Pick} as an exemplar task and examine policy robustness in each condition separately. Further, we extend the analysis to the model-free methods -- PC-PPO and RGBD-PPO. To ensure all models have similar baseline competency in the original setting, we continue to train all methods beyond the steps shown in \figref{fig:01_downstream_tasks} until convergence. All achieve $>$90\% task success rate. We also consider including additional distraction objects as another perturbation. Results are shown in \figref{fig:02_robustness}. We denote point cloud methods in \textcolor{myRed}{red} and RGB-D in \textcolor{myBlue}{blue}.

For viewpoint changes, we see that both \rgbd policies (\rgbdwm and RGBD-PPO) rapidly drop in success rate for minor changes. This effect results in \textbf{0\% success rate} when pitch or yaw change by more than $\pm$0.1 radians (or about 5.7$^{\circ}$). In contrast, the point cloud-based models are robust even to extreme changes, provided the arm remains clearly in view.
Changes in viewpoint do not distort object shapes or relative distance between points. We hypothesize point cloud-based policies remain performant in these instances as they learn to rely on these features rather than absolute positions or object scales. 

For field of view (FoV) changes, we observe that point cloud policies suffer  a minor penalty under different conditions, yet  RGBD policies achieve a \textbf{0\% success rate} for all perturbed settings. Changes to the FoV greatly affect the captured image -- dramatically scaling the image contents as in the example frames. For RGB-D models, this results in significantly out-of-distribution inputs. For point clouds, the geometric relationships between points and their positions relative to the camera do not shift with the FoV changes. 

For lighting changes, we find RGB-D methods to be impacted by irregular lighting (spotlight) or darker scenes, but respond similarly to point cloud models for other conditions. RGB networks are known to be somewhat robust to lighting changes\cite{Hansen2020GeneralizationIR}, but point cloud models meet or exceed them. 
For additional distractor objects, we see no significant difference between point cloud and \rgbd models -- suggesting the robustness exhibited by point clouds may not extend to settings where changes require that the policy performs higher-order relational reasoning with an increased number of objects.
\subsection{\ours{} adapt more quickly than \rgbd counterparts when trained further in viewpoint perturbed environments.}
\label{sec:adapt}
In case of task failures on novel settings, it is desirable to have a model that can be fine-tuned as quickly as possible and not have to learn about the task from scratch. To test if \ours{} can adapt well in such situations, we select 2 conditions from viewpoint (Rel. Pitch (0.4) and Rel. Yaw (-0.6) in Figure \ref{fig:02_robustness}) and one from lighting (Medium in Figure \ref{fig:02_robustness}) where both \ours{} and RGBD-WM performed nearly equally. Starting from the pretrained models, we trained in the new environments until convergence. For viewpoint changes, we observed a significant difference in sample efficiency with \ours{} requiring 32 and 25 episodic interactions compared to 94 and 70 for \rgbd models (each episode consists of 200 timesteps. However, for the lighting perturbation, we found both RGBD-WM and \ours{} took about 100 episodes to reach 100\% task success -- suggesting that \ours{} adapts quickly in geometrically perturbed situations such as viewpoint.

%% file: sections/07_discussion.tex
\begin{figure}[t]
  \centering
  \includegraphics[width=0.75\linewidth]{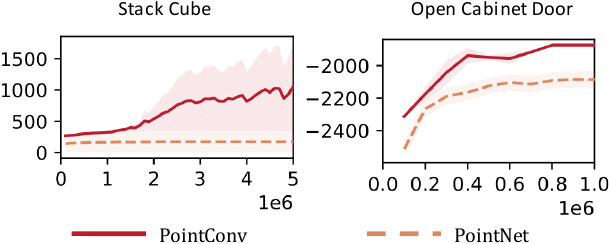}
  \vspace{-5pt}
  \caption{\textit{Comparison of point cloud encoders}: PointConv \cite{wu2019pointconv} consistently shows greater sample efficiency as compared to PointNet \cite{Qi2016PointNetDL} in the \texttt{StackCube} and \texttt{OpenCabinetDoor} tasks.}
  \label{fig:pointnet_comparison}
  \vspace{-11pt}
\end{figure}

Our initial experiments suggest the choice of point cloud encoder is important. \figref{fig:pointnet_comparison} shows a comparison between model-free point cloud-based methods (\ie PC-PPO variants) using encoders based on PointNet \cite{Qi2016PointNetDL} and PointConv \cite{wu2019pointconv} on two tasks. We find significant gains using PointConv -- attributed to PointConv's ability to reason about local points for feature extraction that PointNet lacks.

 While we have shown that \ours{} can be more sample-efficient and robust to visual perturbations, they can be slow to train (wall clock time) when compared to their \rgbd counterparts owing to the point cloud processing. We find \ours{} to be $\sim$2-3.5x slower depending on the number of points in the input. Additionally, we note that the pruning of points needs to be performed on the novel static viewpoint and our system would likely fail with a moving camera.
We hope that the community furthers the research on point cloud models for policy learning and mitigates these limitations.

%% file: sections/08_conclusion.tex
In this work, we presented a novel model-based RL method for partial point cloud observations \ours{}. We demonstrated that this model can result in dramatic sample efficiency on certain tasks, and significant robustness gains over analogous RGB-D models in settings such as viewpoint, field of view and lighting changes. We also showed that the choice of the point cloud network significantly impacts sample efficiency.

%% file: sections/09_acknowledgements.tex
Skand would like to thank Wesley Khademi for his help with PointConv codebase and DMV \& ViRL labmates for providing feedback on an earlier version of the draft. 
This University of Utah effort is supported by DARPA under grant N66001-19-2-4035.
The Oregon State effort is supported in part by the DARPA Machine Common Sense program and ONR awards N00014-22-1-2114, ONR N0014-21-1-2052.

%% file: sections/appendix/01_hyperparameters.tex
We provide the hyper parameters for the (a) world model, (b) policy and (c) general components of \ours{}.
\begin{table}[!h]
\caption{\ours{} hyper parameters. We use the same values for all the tasks except for the the number of input point cloud points ($N$). We vary $N$ depending on the task as reported in \ref{tab: environment_details}.}
\label{tab: hyperparameters}
\begin{center}
\begin{tabular}{ccc}
\toprule
 \textbf{Description} & \textbf{Symbol} & \textbf{Value}
\\
\hline \hline 
\textbf{General} & & \\
\hline \hline 
Number of environments & -- & 1 \\
Number of points (in input point cloud) & $N$ &$\{1024, 2048, 4096\}$ \\
Training Batch size & $B$ & 8 \\
Training sequence length & $T$ & 64\\
Multi-step rollout length & $H$ & 5 \\
\\
\hline \hline 
\textbf{World Model} & & \\
\hline \hline 
Deterministic State & $h_t$ & 256 \\

Stochastic State & $z_t$ & 32 \\
Learning rate & -- & $10^{-4}$ \\
Adam epsilon & -- & $10^{-8}$ \\
Gradient Clipping & -- & 1000 \\
\\
\hline \hline 
\textbf{Actor Critic} & & \\
\hline \hline 
Imagination Horizon & $\mathcal{H}$ & 15 \\
Discount factor & $\gamma$ & 0.997 \\
Return lambda & $\lambda$ & 0.95 \\
Actor Entropy scale & -- & $3 \times 10^{-4}$ \\
Learning rate & -- & $3 \times 10^{-5}$ \\
Adam epsilon & -- & $10^{-5}$ \\
Gradient Clipping & -- & 100 \\

\bottomrule
\end{tabular}
\end{center}
\vskip -0.1in
\end{table}

%% file: sections/appendix/02_environment_details.tex
Here we provide the action space for each of the manipulation and the corresponding number of points chosen for the training of \ours{}.
\begin{table}
\caption{We vary the number of input points depending on the task so as to include maximum possible information in the observation for the agent}
\label{tab: environment_details}
\begin{center}
\begin{tabular}{ccc}
\toprule
 \textbf{Environment} & \textbf{Action dim} & \textbf{Num. Points} \\
\midrule
Lift Cube & 8 & 1024 \\
Clutter Pick & 8 & 1024 \\
Stack Cube & 8 & 1024 \\
Open Cabinet Door & 12 & 2084 \\
Open Cabinet Drawer & 12 & 2084 \\
Move Bucket & 20 & 4096 \\
\bottomrule
\end{tabular}
\end{center}
\end{table}